\documentclass[conference]{IEEEtran}
\IEEEoverridecommandlockouts
\usepackage{cite}
\usepackage{amsmath,amssymb,amsfonts}
\usepackage{graphicx}
\usepackage{textcomp}
\usepackage{xcolor}
\def\BibTeX{{\rm B\kern-.05em{\sc i\kern-.025em b}\kern-.08em
    T\kern-.1667em\lower.7ex\hbox{E}\kern-.125emX}}
    
\usepackage{amsmath,amssymb,amsfonts,amsthm}
\usepackage{bm,bbm}
\usepackage{algorithm}
\usepackage{algpseudocode}
\usepackage{footnote}
\usepackage{comment}
\usepackage{multirow}
\usepackage{epstopdf}
\usepackage{caption,subcaption}
\usepackage{lipsum}
\usepackage{mwe}
\usepackage[normalem]{ulem}
\allowdisplaybreaks
\usepackage{graphicx,lipsum,wrapfig} % text besides figure

\algrenewcommand\algorithmicrequire{\textbf{Input:}}
\algrenewcommand\algorithmicensure{\textbf{Initialize:}}

\begin{document}

\title{Communication Efficient Federated Learning via Ordered ADMM in a Fully Decentralized Setting
}

\author{
\IEEEauthorblockN{Yicheng Chen}
\IEEEauthorblockA{\textit{ECE, Lehigh University} \\
Bethlehem, PA \\
yic917@lehigh.edu}
\and
\IEEEauthorblockN{Rick S. Blum}
\IEEEauthorblockA{\textit{ECE, Lehigh University} \\
Bethlehem, PA \\
rblum@eecs.lehigh.edu}
\and
\IEEEauthorblockN{Brian M. Sadler}
\IEEEauthorblockA{ \textit{Army Research Laboratory} \\
Adelphi, MD \\
brian.m.sadler6.civ@army.mil}
}

\maketitle

\begin{abstract}
The challenge of communication-efficient distributed optimization has attracted attention in recent years. In this paper, a communication efficient algorithm, called ordering-based alternating
direction method of multipliers (OADMM) is devised in a general fully decentralized network setting where a worker can only exchange messages with neighbors. Compared to the classical ADMM, a key feature of OADMM is that transmissions are ordered among workers at each iteration such that a worker with the most informative data broadcasts its local variable to neighbors first, and neighbors who have not transmitted yet can update their local variables based on that received transmission. In OADMM, we prohibit workers from transmitting if their current local variables are not sufficiently different from their previously transmitted value. A variant of OADMM, called SOADMM, is proposed where transmissions are ordered but  transmissions are never stopped for each node at each iteration.   Numerical results demonstrate that given a targeted accuracy, OADMM can significantly reduce the number of communications compared to existing algorithms including ADMM. We also show numerically that SOADMM can accelerate convergence, resulting in communication savings compared to the classical ADMM.
\end{abstract}


\section{Introduction}

In recent years, distributed optimization/learning algorithms have created a great deal of interest,  with particular emphasis on approaches that attempt to optimize a performance criterion employing available data stored at local devices.  The basic idea in distributed learning is to parallelize the computing process across multiple local devices (a.k.a. workers or nodes) to solve the following distributed learning problem
\begin{align}\label{basicprobNSF2020}
\min _{\boldsymbol{\theta} \in \mathbb{R}^{q}} \mathcal{L}(\boldsymbol{\theta}) \quad \text { with } \quad \mathcal{L}(\boldsymbol{\theta}) \buildrel \Delta \over=\sum_{m \in \mathcal{M}} \mathcal{L}_{m}(\boldsymbol{\theta})
\end{align}
where $\boldsymbol{\theta} \in \mathbb{R}^{q}$ is the model parameter vector, $\mathcal{L}(\boldsymbol{\theta})$ is the global objective function, and $\mathcal{L}_{m}(\boldsymbol{\theta})$ is the local function for worker $m \in \mathcal{M}=\{1,2,...,M\}$ that is often defined as $\mathcal{L}_{m}(\boldsymbol{\theta})\buildrel \Delta \over=\sum_{n=1}^{N_m}\ell(\boldsymbol{\theta};\boldsymbol{x}_n,y_n) $  where $\ell(\boldsymbol{\theta};\boldsymbol{x}_n,y_n)$ is the loss function for $n=1,2,...,N_m$, $\boldsymbol{x}_n$ is the $n$-th feature vector and $y_n$ is the corresponding label. One common approach to solve (\ref{basicprobNSF2020}) is a {\em worker-server} architecture where the server regularly broadcasts the model parameter to all workers, the workers calculate their local gradients based on the received model parameter and send them back to the central server, and finally the server aggregates the gradients received from all workers to update the model parameter. However, this architecture may be impractical in some cases and the server becomes a potential single point of failure \cite{xin2020decentralized, roy2019braintorrent}. 
%due to single point failure and a lack of availability of a server node which is not battery powered, 
%limited transmission power of the server, 
%this architecture might not be desirable or available to consider 

In this paper, we consider a more general and robust distributed network architecture where each worker exchanges information with its neighbors. Each node then updates its model parameter based on the information from its neighbors. Although this fully decentralized architecture enjoys better scalability, a very large number of communications among all the nodes are typically required to solve (\ref{basicprobNSF2020}).  
In a wireless communications setting a large number of transmissions is undesirable due to energy consumption. 
%If the nodes are battery powered, common in a wireless communication setting, a large number of communications can drain these batteries.  
Thus, reducing the number of communications 
is highly desirable. 
In addition, for typical computing hardware,
the latency to send data over a network connection is much larger than that for accessing data in its own main memory \cite{smith2017cocoa}.  
Thus, the communications can be a significant bottleneck that lengthens the overall time to complete an algorithm. 
The goal of this paper is to develop a class of communication-efficient distributed learning algorithms to accurately solve (\ref{basicprobNSF2020}) while reducing the communications in a fully decentralized  architecture setting.

A straightforward method to enhance communication efficiency is to accelerate convergence, which also tends to reduce communications. Some researchers have proposed one-shot parameter averaging methods in \cite{zhang2013communication, mcdonald2009efficient, mcmahan2016federated} to find the optimal minimizer using only one iteration, which might not be stable in some cases \cite{shamir2014communication}. Some popular primal-dual methods \cite{smith2017cocoa, duchi2011dual, scaman2018optimal, he2018cola} are also shown to be efficient in federated learning where the primal solution is obtained by efficiently solving the dual problem. 
Another alternative uses alternating direction method of multipliers (ADMM) algorithms 
%to gain 
%the desired property of parallelization
where the workers alternate between computing the dual variables in a distributed way, and solving augmented Lagrangian problems based on their own data \cite{shamir2014communication, boyd2011distributed, hong2017linear, deng2016global, zhang2014asynchronous, makhdoumi2017convergence, shi2014linear, deng2017parallel, wang2019global}. 
These algorithms are very effective, and so it is of interest to devise versions of ADMM that require fewer transmissions to achieve a given accuracy.
%per iterations.
%However, these algorithms, do not consider reducing the number of transmissions per iteration.

%One interesting approach called 
{\em Censoring} has been shown to
be an effective method to improve efficiency, where workers only transmit  highly informative data and thereby reduce the number of transmissions \cite{rago1996censoring}.
In \cite{liu2019communication} ADMM with censoring (censoring-ADMM) was proposed, that restricts each worker from transmitting its local message to neighbors if it is not sufficiently different from the previously transmitted one, and neighbors can use the past received value from that worker to approximate the current one. The sufficiency of the difference needed for transmission is adaptively determinted using a predefined censoring function. However, there are some aspects of ADMM and transmission reduction 
%distributed data processing strategies 
that have not yet been explored.
%exploited by the censoring-ADMM scheme. 
For example,  in censoring-ADMM, each worker independently decides whether to transmit or not without interacting with other workers. Our proposed algorithm gains substantial performance enhancement by exploiting how transmissions can be arranged in time and processed such that workers can collaboratively decide what is the most informative data to transmit to their neighbors.

% based on properly ordered transmissions such that workers can collaboratively decide what is the most informative data to transmit to their neighbors.  more complicated but efficient algorithm, called Alternating Direction Method of Multipliers (ADMM)

Our proposed algorithms
%presented later in this paper, 
%employs our own newly developed algorithms which 
%build on an idea we developed called 
employ {\em ordering} such that at each iteration nodes with more informative data transmit their messages sooner 
%first to their neighbors 
\cite{blum2008energy}. To the best of our knowledge, ordered transmissions have not been applied to 
%we have not seen any other work that is aimed at applying ordering to 
federated learning in a completely distributed setting.  
%In ordering, each worker will decide, in a distributed way, how informative is its own data, such that the worker with the most informative data will transmit first while the workers with less informative data will transmit later.  
Some extensions to the work in \cite{blum2008energy} have been
developed, including the application of ordering to quickest change detection in sensor networks \cite{chen2021ordering}, nearest-neighbor learning \cite{marano2013nearest}, and ordered gradient descent (GD)  in a  worker-server architecture setting \cite{chen2020ordered}. 

%In this paper, we provide the first use of ordering to ADMM, called OADMM, with the
In this paper we develop ordered ADMM (OADMM), with the goal of finding the model parameter $\boldsymbol{\theta}$ that minimizes (\ref{basicprobNSF2020}) while reducing the communications among the nodes with respect to those required for ADMM {for a given network topology}. Similar to censoring-ADMM, a node is allowed to transmit its local variable to its neighbors only if its current local variable is significantly different from its previously transmitted one. {The key novelty of OADMM is that for each iteration (i) the transmission-ordering strategy is employed, and (ii) 
every node will update its local estimate using the information received prior to its own transmission time. (The transmission time for each node is fixed at the beginning of each iteration, and the mid-iteration update in (ii) does not alter the ordering.) }
%from the nodes that have already transmitted during the current iteration.}  
%where transmissions are ordered  so that a node with less informative data can update its local variable prior to its own broadcast to its neighbors, 
%before broadcasting its local variables after receiving messages from its neighbors who have more informative data. 
Thus, the most informative data is used by all neighbors who have not yet transmitted, and numerical results show that ordering and processing in this way yields significant communication savings. 
A cutoff time threshold is employed such that for each iteration, only a subset of the nodes (the most informative nodes based on ordering) will transmit.
Furthermore, when we allow all nodes to transmit at each iteration (by not employing the cutoff time threshold), OADMM reduces to a variant of the classical ADMM, called SOADMM, which is also shown to have a better convergence behavior than ADMM thanks to ordering.

\section{Classical ADMM}

Before introducing OADMM, we first establish the model and briefly review ADMM in the decentralized setting. 
%introduce a general connected network architecture employed in this paper and review the classical ADMM in this network architecture setting.
The network is characterized by an underlying undirected graph  $\mathcal{G}=\{\mathcal{M},\mathcal{E}\}$, where $\mathcal{M}=\{1,2,...,M\}$ is a set of nodes and $\mathcal{E}$ denotes the set of undirected edges of the graph. Note that in this paper, the terms ``worker'' and ``node'' are used interchangeably. Node $m$ and node $m'$ are called neighbors if the edge $(m,m')\in\mathcal{E}$. We denote the set of neighbors of node $m$ as $\mathcal{N}_m$ with $d_m=|\mathcal{N}_m|$.

To solve (\ref{basicprobNSF2020}) with classical ADMM, at each iteration $k$ each node $m$ broadcasts its local solution ${\boldsymbol{\theta}}_m^k$ to its neighbors and then updates its local estimate using %according to the information received from its neighbors and its own local function \cite{zhang2012efficient}.
\begin{align}\label{classicalADMMprimal}
&{\boldsymbol{\theta}}_m^k =\arg \min _{{\boldsymbol{\theta}}_m} \mathcal{L}_m\left({\boldsymbol{\theta}}_m\right)\notag\\
&\quad+\left\langle {\boldsymbol{\theta}}_m, \lambda_{m}^{k-1}-\alpha \sum_{m' \in \mathcal{N}_{m}}\left({\boldsymbol{\theta}}_m^{k-1}+ {\boldsymbol{\theta}}_{m'}^{k-1}\right)\right\rangle+ \alpha d_{ m}\left\|{\boldsymbol{\theta}}_m\right\|^{2}.
\end{align}
The local dual variable $\lambda_{m}^{k}$ is then computed using
\begin{align}\label{classicalADMMdual}
\lambda_{m}^{k}=\lambda_{ m}^{k-1}+\alpha \sum_{m' \in \mathcal{N}_{ m}}\left({\boldsymbol{\theta}}_{m}^{k}-{\boldsymbol{\theta}}_{ m'}^{k}\right).
\end{align}
Each ADMM iteration typically requires
%It is worth mentioning that
$\sum_{m=1}^Md_m$  communications, and this can be significantly reduced using OADMM as described next.
%are typically needed for each iteration of the classical ADMM, which can be significantly reduced by using the following proposed OADMM algorithm.

\section{Ordering-based ADMM}

In this section,  we consider applying ordering to ADMM in a general network architecture setting where each worker communicates with a subset of the other workers in the network.  

%   In censoring-based ADMM \cite{liu2019communication}, a node does not necessarily broadcast at every iteration, unlike in classical ADMM. It only broadcasts its local solution to neighbors if the local solution is sufficiently different from the previously transmitted
% one. Although censoring-based ADMM can help us reduce the number of broadcasts, there are some aspects of distributed networks that have not yet been exploited by it and it is possible to significantly reduce the number of broadcasts in cases of interest.   To demonstrate this, an ordering-based ADMM (OADMM) method which exploits these unexploited aspects is developed next.  

Like classical ADMM, OADMM is synchronous such that all nodes start each iteration at the same time. Additionally, we denote {$\Delta_k$} as the cutoff  time  threshold of iteration $k$ and denote $t_k$ as the starting time of iteration $k$. It follows that $t_{k+1}=t_k + \Delta_k$ for $k=1,2,...,K$ where $K$ is the total number of iterations. At iteration $k$ of OADMM, node $m$ not only computes the primal variable ${\boldsymbol{\theta}}_m^k$ and the dual variable $\lambda_m^k$ but also stores its last dual variable $\lambda_m^{k-1}$ and its state variable $\hat{\boldsymbol{\theta}}_m^{k-1}$ that describes its most recent  primal variable broadcast prior to iteration $k$. 
Similar to classical ADMM, besides these variables, the state variables $\hat{\boldsymbol{\theta}}_{m'}^{k-1}$ for all its neighbors $m'\in\mathcal{N}_m$ are also recorded at node $m$. 

An important feature of OADMM is that nodes will transmit in order with most informative first. 
%, to send more informative information first. 
Specifically, at $t_k$ (beginning of iteration $k$), each node will calculate the {initial} primal variable ${ \tilde{{\boldsymbol{\theta}}}_m^k}$ using  
\begin{align}\label{firstupdateOADMM}
&{ \tilde{{\boldsymbol{\theta}}}_m^k} =\arg \min _{{\boldsymbol{\theta}}_m} \mathcal{L}_m\left({\boldsymbol{\theta}}_m\right)\notag\\
&\quad+\left\langle {\boldsymbol{\theta}}_m, \lambda_{m}^{k-1}-\alpha \sum_{m' \in \mathcal{N}_{m}}\left(\hat{\boldsymbol{\theta}}_m^{k-1}+ \hat{\boldsymbol{\theta}}_{m'}^{k-1}\right)\right\rangle+ \alpha d_{ m}\left\|{\boldsymbol{\theta}}_m\right\|^{2}
\end{align}
where $\alpha$ is the step size and  $d_m=|\mathcal{N}_m|$ is the number of neighbors of node $m$. Then, each node $m=1,\dots,M$ sets a timer and waits $\tau/(c_0+\|{ \tilde{{\boldsymbol{\theta}}}_m^k}-\hat{\boldsymbol{\theta}}_m^{k-1}\|)$ seconds to broadcast to its neighbors.
%after the timer started.  
Here $\tau$ is a positive number that is set as small as the physical system will allow and $c_0$ is a predefined constant. The nodes with the largest $\|{ \tilde{{\boldsymbol{\theta}}}_m^k}-\hat{\boldsymbol{\theta}}_m^{k-1}\|$ transmit first and the nodes with  smaller values transmit later. 
{If the transmit time exceeds the cutoff, then these nodes do not transmit. }
%if they are allowed to transmit. 
{We denote the set of neighbors of node $m$ who have broadcasted before and after node $m$ at iteration $k$ as $\mathcal{N}_{m,k}''$ and $\mathcal{N}_{m,k}'$, respectively. Note that $\mathcal{N}_{m}=\mathcal{N}_{m,k}''\bigcup\mathcal{N}_{m,k}'$. Just before it transmits,}
{node $m$ performs an update using the new information from its neighbors, given by}
{\begin{align}\label{newupdateADMM}
&{\boldsymbol{\theta}}_{m}^k =\arg \min _{{\boldsymbol{\theta}}_{m}} \mathcal{L}_{m}\left({\boldsymbol{\theta}}_{ m}\right)\notag\\
&\qquad+
\left\langle {\boldsymbol{\theta}}_{m}, \lambda_{m}^{k-1}-\alpha \sum_{\substack{ m' \in \mathcal{N}_{m,k}'}}\left(\tilde{{\boldsymbol{\theta}}}_m^k+ \hat{\boldsymbol{\theta}}_{m'}^{k-1}  \right)\right.\notag\\
&\qquad \left.-\alpha \sum_{\substack{ m'' \in \mathcal{N}_{m,k}''}}\left(\tilde{{\boldsymbol{\theta}}}_m^k+  {\boldsymbol{\theta}}_{m''}^k \right) \right\rangle+ \alpha d_{m}\left\|{\boldsymbol{\theta}}_{m}\right\|^{2}.
\end{align}}

{Immediately after broadcasting, node $m$ will update its own state variable $\hat{\boldsymbol{\theta}}_m^{k}={\boldsymbol{\theta}}_{m}^k$.}
During iteration $k$, if node $m$ for all $m=1,2,...,M$ receives the primal variable ${\boldsymbol{\theta}}_{m'}^k$ from its neighbor $m'\in\mathcal{N}_m$, then node $m$ will locally update the information about its neighbor by setting $\hat{\boldsymbol{\theta}}_{m'}^{k}={\boldsymbol{\theta}}_{m'}^k$. 
If such an update does not occur, then the value of this variable at node $m$ maintains its previous value $\hat{\boldsymbol{\theta}}_{m'}^{k}=\hat{\boldsymbol{\theta}}_{m'}^{k-1}$. 
Any nodes who have not yet transmitted prior to time  ($t_k+{\Delta_k}$) will not transmit  
during iteration $k$.  At time ($t_k+{\Delta_k}$), node $m$ will compute its dual variable $\lambda_m^k$ (for $m=1,\ldots,M$) using
\begin{align}\label{updatedualvariable}
\lambda_{m}^{k}=\lambda_{ m}^{k-1}+\alpha \sum_{m' \in \mathcal{N}_{ m}}\left(\hat{\boldsymbol{\theta}}_{m}^{k}-\hat{\boldsymbol{\theta}}_{ m'}^{k}\right).
\end{align}

OADMM is summarized in Algorithm \ref{OADMM}. If all transmission propagation delays are known and timing is synchronized, one can schedule all transmissions 
%of its neighbors back to a given node
so they arrive in the correct order. However, even with 
%inaccurate estimates of propagation delays or 
imperfect  synchronization 
%since the node receives the values to be ordered, 
a node can put them back in order correctly as long as the node waits a short period relative to the uncertainty; see \cite{blum2011ordering}. 

{Setting the cutoff  time  threshold to be ${\Delta_k}=\tau/c_0$ 
{effectively eliminates the cutoff so that every node transmits in each iteration. We call this variation SOADMM.
\textcolor[rgb]{0.00,0.00,0.00}{Compared to the classical ADMM, numerical results in Section \ref{numericalOADMM} show that 
both OADMM and SOADMM can save communications for a given accuracy, and SOADMM can 
accelerate convergence.} }
%This ability to accelerate convergence opens new avenues of investigation. 
}
%so that no transmissions are stopped for each node at each iteration in Algorithm \ref{OADMM}} defines a new algorithm we call SOADMM. 

\begin{savenotes}
\begin{algorithm}
\caption{OADMM. }\label{OADMM}
\begin{algorithmic}[1]
\State \textcolor[rgb]{0.00,0.00,0.00}{Assume syn. and known transmission delays (see first new paragraph
after (\ref{updatedualvariable}) for relaxing.).}
\Require{step size $\alpha$, constants $c_0>0$, $c_1>0$, $0<\rho<1$, and $\tau>0$.}
\Ensure{\{${\boldsymbol{\theta}}_m^0=0$, $\lambda_m^0=0$, $\hat{\boldsymbol{\theta}}_m^0=0$, $\forall m$\}}
\For{$k=1, 2,..., K$}
\State Each node $m$ updates its {initial} primal variable ${ \tilde{{\boldsymbol{\theta}}}_m^k}$ via (\ref{firstupdateOADMM}) at the starting time $t_k$ of iteration $k$.
\State Each node $m$ will determine a time $
t_{m,k}=t_k +  
\tau/(c_0+\|\textcolor[rgb]{0.00,0.00,0.00}{ \tilde{{\boldsymbol{\theta}}}_m^k}-\hat{\boldsymbol{\theta}}_m^{k-1}\|)$ to broadcast.
\State Order nodes $m_1,...,m_M$ using $t_k\le t_{m_1,k}\le t_{m_2,k}\le...\le t_{m_M,k}$.
\State Set $i=1$ and  {${\Delta_k}=\tau/{(c_0+c_1\rho^k)}$}.
\While{$i\le M$}
\If{$t_{m_i,k}\le t_k+{\Delta_k}$}
% \If{node $m_i$ receives ${\boldsymbol{\theta}}_{m_{i}''}^k$ from $m_{i}''\in\mathcal{N}_{m_i}$ before time $t_{m_i,k}$}
% \State Node $m_i$ updates  ${\boldsymbol{\theta}}_{m_i}^k$ again via (\ref{newupdateADMM}) .
% \EndIf
\State \textcolor[rgb]{0.00,0.00,0.00}{At time $t_{m_i,k}$, node $m_i$ updates  ${\boldsymbol{\theta}}_{m_i}^k$ via (\ref{newupdateADMM}) .}
\State  At time $t_{m_i,k} $, node $ m_i $ broadcasts ${\boldsymbol{\theta}}_{m_i}^k$ to  \textcolor[rgb]{0.00,0.00,0.00}{all of its neighbors and updates its state variable $\hat{\boldsymbol{\theta}}_{m_i}^{k}={\boldsymbol{\theta}}_{m_i}^k$ immediately after broadcasting}.
\State \textcolor[rgb]{0.00,0.00,0.00}{At time $t_{m_i,k} $, \textcolor[rgb]{0.00,0.00,0.00}{all neighbors of node $m_i$}  update the information about node $m_i$  as $\hat{\boldsymbol{\theta}}_{m_i}^k = {\boldsymbol{\theta}}_{m_i}^k$.}
\Else
\State \textcolor[rgb]{0.00,0.00,0.00}{At time $t_k+{\Delta_k}$, node $m_i$ updates its state variable $\hat{\boldsymbol{\theta}}_{m_i}^{k}=\hat{\boldsymbol{\theta}}_{m_i}^{k-1}$.}
\State \textcolor[rgb]{0.00,0.00,0.00}{At time $t_k\!+\!{\Delta_k}$, \textcolor[rgb]{0.00,0.00,0.00}{all neighbors of node $m_i$} update information about node $m_i$  as $\hat{\boldsymbol{\theta}}_{m_i}^k\!\!=\! \hat{\boldsymbol{\theta}}_{m_i}^{k-1}$\!.}
\EndIf
\State At time $t_k+{\Delta_k}$, node $m_i$ updates its local dual variable $\lambda_{m_i}^k$ via (\ref{updatedualvariable}).
\State Set $i=i+1$.
\EndWhile
\EndFor
\end{algorithmic}
\end{algorithm}
\end{savenotes}

\addtolength{\topmargin}{0.02in}

\section{Numerical Results}\label{numericalOADMM}
Consider the linear regression task studied in  \cite{liu2019communication}, with the local function at worker $m$ in (\ref{basicprobNSF2020}) being
\begin{align}\label{locallinearregression}
\mathcal{L}_{m}(\boldsymbol{\theta})\buildrel \Delta \over=\frac{1}{2}\sum_{n=1}^{{N}_{m}}\left(y_{n}-\mathbf{x}_{n}^{\top} \boldsymbol{\theta}\right)^{2}
\end{align}
{where $y_n$ is the $n$-th label and $\mathbf{x}_{n}$ is the corresponding feature vector. Let ${\boldsymbol{\theta}}^*$ denote the optimal solution. 
\textcolor[rgb]{0.00,0.00,0.00}{Each scalar in the vector  $( ({\boldsymbol{\theta}}^*)^{\top}, \mathbf{x}_{1}^{\top},\ldots,\mathbf{x}_{N_m}^{\top})$  is chosen uniformly from the set $[0.1,0.2,...,1]$ and we obtain $y_n$ by using $y_n=\mathbf{x}_{n}^{\top}{\boldsymbol{\theta}}^*$.}}
We set $M=50$ and $N_m=3$ for all $m=1,...,M$.
We compare the performance of OADMM with classical ADMM, censoring-based ADMM \cite{liu2019communication}, and SOADMM using the same setting/problem as in \cite{liu2019communication} to provide a favorable environment for censoring-based ADMM. As in \cite{liu2019communication}, the network has $10\%$ of all possible edges randomly chosen to be connected. %is employed. 
The step size $\alpha=0.4$ is used for all four algorithms. 
%classical ADMM, OADMM, SOADMM, and censoring-based ADMM. 
\textcolor[rgb]{0.00,0.00,0.00}{We employ the optimized stopping threshold $ c_1\rho^k $ ($c_1=5$, $\rho=0.87$) in censoring-based ADMM} \cite{liu2019communication}.  {We allow workers to transmit at iteration $k$ in OADMM if  $\|\tilde{{\boldsymbol{\theta}}}_m^k-\hat{\boldsymbol{\theta}}_m^{k-1}\|\ge c_1\rho^k $  and  let all workers transmit in SOADMM}.

\begin{figure}
\centering
\begin{subfigure}{.5\textwidth}
  \centering
  \includegraphics[width=1.\linewidth]{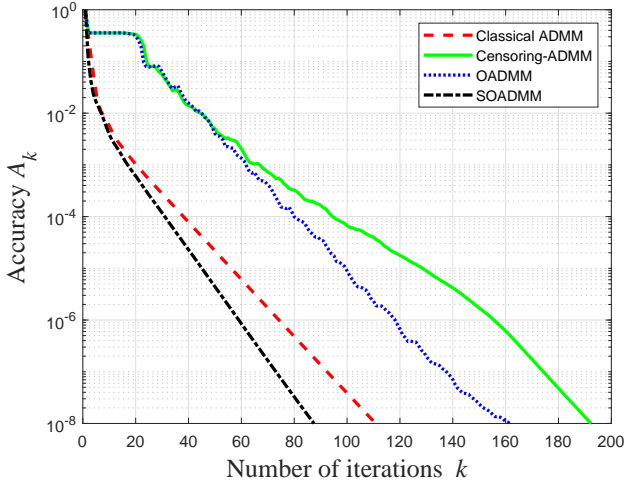}
  \caption{Accuracy versus {number of iterations}}
  \label{ADfig:sub1}
\end{subfigure}%
\hfill
\begin{subfigure}{.5\textwidth}
  \centering
  \includegraphics[width=1.\linewidth]{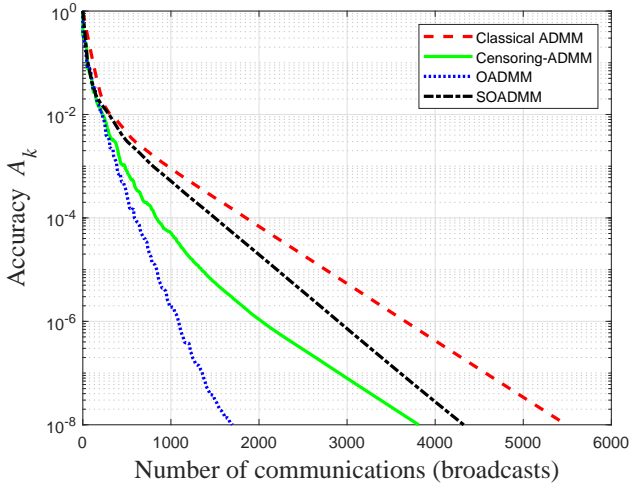}
  \caption{Accuracy versus {number of} transmissions}
  \label{ADfig:sub2}
\end{subfigure}
\caption{Accuracy versus (a) the number of iterations, and (b) the total number of communications, for the linear regression example with a random network topology.}
\label{fig:OADMM}
\end{figure}

Let ${\boldsymbol{\theta}}_m^0$ denote the initial value of ${\boldsymbol{\theta}}_m^k$. Define the accuracy at iteration $k$ as
${\it A}_k=(\sum_{m=1}^M\|{\boldsymbol{\theta}}_m^k-{\boldsymbol{\theta}}^*\|^2)/(\sum_{m=1}^M\|{\boldsymbol{\theta}}_m^0-{\boldsymbol{\theta}}^*\|^2)$. The accuracy ${\it A}_k$ is plotted versus the number of iterations $k$ in Fig. \ref{fig:OADMM}(a).  By counting the total number of communication transmissions during each iteration $k$, the accuracy 
%(still denoted by ${\it A}_k$ for convenience) 
is plotted in Fig. \ref{fig:OADMM}(b) versus the 
total number of transmissions by all the nodes up to the $k$-th iteration.  Fig. \ref{fig:OADMM} illustrates that OADMM is able to significantly reduce the total number of transmissions when compared to ADMM, censoring-based ADMM \cite{liu2019communication}, and SOADMM.  Given a target accuracy of $10^{-8}$, OADMM can save about $70\%$ of the total transmissions compared to classical ADMM. 
{Comparing Fig. \ref{fig:OADMM}(a) and \ref{fig:OADMM}(b), to reach the same accuracy, SOADMM requires fewer iterations whereas OADMM requires fewer total transmissions. }
%{Compared to classical ADMM, Fig. \ref{fig:OADMM} indicates that SOADMM requires a smaller number of iterations which shows the potential for ordering to accelerate the optimization process.} 

% \textcolor[rgb]{1.00,0.00,0.00}{Next we consider the total number of communication broadcasts needed for the classical ADMM and OADMM to achieve a given accuracy $10^{-8}$ versus the number of nodes in a random network where $15\%$ of all possible edges are randomly chosen to be connected. All other parameters are set to be the same as Fig. \ref{fig:OADMM}. The result in Fig. \ref{Comms_versus_NumNodes} shows that OADMM saves a significant number of communications while achieving a comparable accuracy compared to the classical ADMM. It also indicates that more communications are typically needed when a network has more nodes. As the number of nodes becomes larger, Fig. \ref{Comms_versus_NumNodes} shows that the communication savings of OADMM over the classical ADMM tend to be larger, which indicates OADMM might be especially helpful for a relatively large network.}

\begin{figure}[!t]
\centering
\includegraphics[width=3.5in]{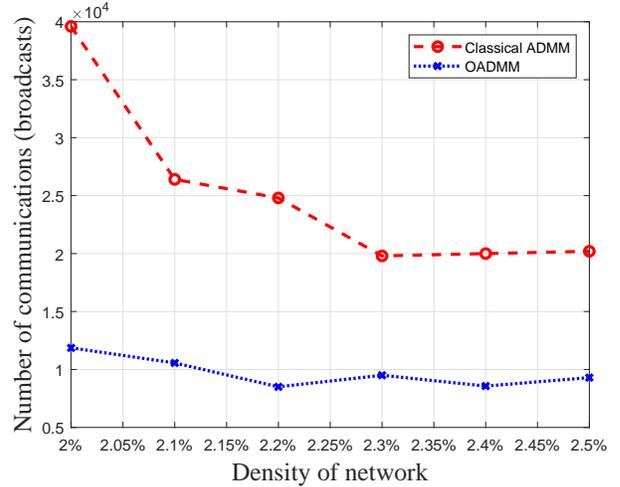}
\caption{Total number of transmissions to achieve $10^{-8}$ accuracy versus the network density (percentage of all possible edges) for linear regression with a random network topology ($M=200$).}
\label{Comms_versus_density}
\end{figure}

{In Fig. \ref{Comms_versus_density}, we plot the number of communications needed to achieve accuracy $10^{-8}$ for OADMM and ADMM versus the network density. {Here, density is defined as the average number of neighbors connected, divided by the network size $M$. In this example we fix $M=200$, and} all other parameters are set to be the same as 
the previous example.
%Fig. \ref{fig:OADMM}. 
Fig. \ref{Comms_versus_density} shows that OADMM achieves the same accuracy as ADMM with many fewer total transmissions.}
%outperforms the classical ADMM in terms of the number of communications needed. 

{ADMM requires every node to transmit at each iteration, 
so for ADMM the total number of transmissions scales directly with the number of iterations needed to achieve the desired accuracy.
\textcolor[rgb]{0.00,0.00,0.00}{Note that when ADMM is applied to linear regression, 
%a relatively sparse network can converge more rapidly 
%nearly as fast as a fully connected network, assuming 
it is possible to optimize the stepsize for different network densities   
%can be found that depends on network density
\cite{shi2014linear}.  Here we continue to use the stepsize from Fig. \ref{fig:OADMM} which was suggested in \cite{liu2019communication}. 
%, and consequently ADMM convergence is sensitive to the network density. 
Fig. \ref{Comms_versus_density} shows that, for this range of densities, OADMM is less sensitive to the changing network density and generally saves more than half the total number of transmissions.
}}
%ince the number of communications per iteration is fixed for ADMM,  Fig. \ref{Comms_versus_density} also shows that a higher network density leads to faster convergence (fewer communications) when we vary the density within a certain range, e.g., in this case $[2\%,2.3\%]$.  Apparently, adding edges in a sparse network is helpful in this case, see \cite{shi2014linear} for further results on this issue for ADMM. %However, we also observe that adding more edges for ADMM and OADMM for very large densities does not necessarily reduce the number of communications, e.g., the curve for ADMM becomes flat when the density is larger than $2.3\%$. This is consistent with what has been shown in \cite{shi2014linear}. 

% \begin{figure}[!t]
% \centering
% \includegraphics[width=3.5in]{Comms_versus_NumNodes.eps}
% \caption{Number of communications (broadcasts) to achieve $10^{-8}$ accuracy versus the number of nodes in linear regression with the random network topology ($15\%$ edges are connected).}
% \label{Comms_versus_NumNodes}
% \end{figure}

\section{CONCLUSIONS}
{
We devised OADMM, a communication-efficient version of ADMM, to accurately solve fully distributed learning tasks. OADMM is able to significantly reduce the number of communications between workers compared to ADMM, saving energy and potentially reducing the time to solution.
Transmissions are ordered at each iteration such that a worker with more informative data broadcasts its local variable sooner, and its neighbors take advantage of this received information to update their local variables before they broadcast during the current iteration. Thus, compared with ADMM, OADMM requires that some nodes perform an additional computation within each iteration. Similar to censoring, a worker is not allowed to broadcast if its current local variable is not sufficiently different from its previously transmitted one, and its neighbors use the last received value from that worker to approximate the current one. }

{
OADMM employs a cutoff time, while a variant called SOADMM allows all workers to transmit at each iteration.
Compared with ADMM, SOADMM accelerates convergence (requires fewer total iterations), whereas OADMM may require more iterations but use fewer total transmissions.  
Numerical results show that both OADMM and SOADMM require many fewer total transmissions than ADMM to achieve the same accuracy. 
It is of interest in future work to show theoretical convergence properties of OADMM and SOADMM. 
It may also be possible to combine OADMM with quantization and/or sparsification techniques to further reduce the communications load in bits.  Robust operation is also of practical interest, to relax the network timing requirements. \textcolor[rgb]{0.00,0.00,0.00}{Asynchronous operation is also of great interest.}
}

\bibliographystyle{IEEEtran}
\bibliography{refs}

\end{document}